\documentclass[fleqn,10pt,dvipsnames]{wlscirep}
\usepackage[utf8]{inputenc}
\usepackage[T1]{fontenc}

\usepackage{csquotes}

\title{Adaptive Partial Scanning Transmission Electron Microscopy with Reinforcement Learning}

\author[1,a]{Jeffrey M. Ede}
\affil[1]{University of Warwick, Department of Physics, Coventry, CV4 7AL, UK}
\affil[a]{j.m.ede@warwick.ac.uk}

\usepackage{microtype}

\newcommand*\linkcolours{CornflowerBlue}

\usepackage{times}
\usepackage{graphicx}
\usepackage{amssymb}
\usepackage{gensymb}
\usepackage{amsmath}
\usepackage{breakurl}

\usepackage{url,hyperref}
\hypersetup{
colorlinks,
linkcolor=\linkcolours,
citecolor=\linkcolours,
filecolor=\linkcolours,
urlcolor=\linkcolours
}

\usepackage{algorithm}
\usepackage{algorithmic}

\usepackage{mathtools, cuted}

\usepackage{tabularx}
\usepackage{multirow}

\usepackage{float}

\newcolumntype{Y}{>{\centering\arraybackslash}X}

\usepackage[]{placeins}

\usepackage{pgf}

\usepackage{tikz}

\begin{abstract}
Compressed sensing can decrease scanning transmission electron microscopy electron dose and scan time with minimal information loss. Traditionally, sparse scans used in compressed sensing sample a static set of probing locations. However, dynamic scans that adapt to specimens are expected to be able to match or surpass the performance of static scans as static scans are a subset of possible dynamic scans. Thus, we present a prototype for a contiguous sparse scan system that piecewise adapts scan paths to specimens as they are scanned. Sampling directions for scan segments are chosen by a recurrent neural network based on previously observed scan segments. The recurrent neural network is trained by reinforcement learning to cooperate with a feedforward convolutional neural network that completes the sparse scans. This paper presents our learning policy, experiments, and example partial scans, and discusses future research directions. Source code, pretrained models, and training data is openly accessible at \url{https://github.com/Jeffrey-Ede/adaptive-scans}.
\end{abstract}

\begin{document}

\flushbottom
\maketitle

\section{Introduction}

Most scan systems sample signals at sequences of discrete probing locations. Examples include atomic force microscopy\cite{krull2020artificial, rugar1990atomic}, computerized axial tomography\cite{new1974computerized, heymsfield1979accurate}, electron backscatter diffraction\cite{schwartz2009electron}, scanning electron microscopy\cite{vernon2000scanning}, scanning Raman spectroscopy\cite{keren2008noninvasive}, scanning transmission electron microscopy\cite{tong2018scanning} (STEM) and X-ray diffraction spectroscopy\cite{scarborough2017dynamic}. In STEM, the high current density of electron probes produces radiation damage in many materials, limiting the range and types of investigations that can be performed\cite{Hujsak_2016, Egerton_2004}. In addition, most STEM signals are oversampled\cite{ede2020warwick} to ease visual inspection and decrease sub-Nyquist artefacts\cite{amidror2015sub}. As a result, a variety of compressed sensing\cite{binev2012compressed} algorithms have been developed to enable decreased STEM probing\cite{preprint+ede2020review}. In this paper, we introduce a new approach to STEM compressed sensing where a scan system is trained to piecewise adapt partial scans\cite{preprint+ede2019partial} to specimens by deep reinforcement learning\cite{li2017deep} (RL).

Established compressed sensing strategies include random sampling\cite{hwang2017towards, hujsak2016suppressing, anderson2013sparse}, uniformly spaced sampling\cite{fang2019deep, de2019resolution, preprint+ede2019deep, hujsak2016suppressing}, sampling based on a model of a sample\cite{mueller2011selection, wang2009variable}, partials scans with fixed paths\cite{preprint+ede2019partial}, dynamic sampling to minimize entropy\cite{ji2008bayesian, seeger2008compressed, braun2015info, carson2012communications} and dynamic sampling based on supervised learning\cite{godaliyadda2017framework}. Complete signals can be extrapolated from partial scans by an infilling algorithm, estimating their fast Fourier transforms\cite{ermeydan2018sparse} or inferred by an artificial neural network\cite{preprint+ede2019partial, preprint+ede2019deep} (ANN). In general, the best sampling strategy varies for different specimens. For example, uniformly spaced sampling is often better than spiral paths for oversampled STEM images\cite{preprint+ede2019partial}. However, sampling strategies designed by humans usually have limited ability to leverage an understanding of physics to optimize sampling. As proposed by our earlier work\cite{preprint+ede2019partial}, we have therefore developed ANNs to dynamically adapt scan paths to specimens. Expected performance of dynamic scans can always match or surpass expected performance of static scans as static scan paths are a special case of dynamic scan paths and performance varies for different static scan paths\cite{preprint+ede2019partial}.

Exploration of STEM specimens is a finite-horizon partially observed Markov decision process\cite{saldi2019asymptotic, jaakkola1995reinforcement} (POMDP) with sparse losses: A partial scan can be constructed from path segments sampled at each step of the POMDP and a loss can be based on the quality of an scan completion generated from the partial scan with an ANN. Most scan systems support custom scan paths or can be augmented with a field programmable gate array\cite{sang2017dynamic, Sang2017a} (FPGA) to support custom scan paths. However, there is a delay before a scan system can execute or is ready to receive a new command. Total latency can be reduced by using both fewer and larger steps, and decreasing steps may also reduce distortions due to cumulative errors in probing positions\cite{sang2017dynamic} after commands are executed. Command execution can also be delayed by ANN inference. However, inference delay can be minimized by using a computationally lightweight ANN and inferring future commands while previous commands are executing.

Markov decision processes (MDPs) can be optimized by recurrent neural networks (RNNs) based on long short-term memory\cite{hochreiter1997long, olah2015understanding} (LSTM), gated recurrent unit\cite{cho2014learning} (GRU), or other cells\cite{weiss2018practical, jozefowicz2015empirical, bayer2009evolving}. LSTMs and GRUs are popular as they solve the vanishing gradient problem\cite{pascanu2013difficulty} and have consistently high performance\cite{jozefowicz2015empirical}. Small RNNs are computationally inexpensive and are often applied to MDPs as they can learn to extract and remember state information to inform future decisions. To solve dynamic graphs, an RNN can be augmented with dynamic external memory to create a differentiable neural computer\cite{graves2016hybrid} (DNC). To optimize a MDP, a discounted future loss, $Q_t$, at step $t$ in a MDP with $T$ steps can be calculated from step losses, $L_t$, with Bellman's equation,
\begin{equation}
    Q_t = \sum\limits_{t'=t}^T \gamma^{t'-t} L_{t'} \,,
\end{equation}
where $\gamma \in [0,1)$ discounts future step losses. Equations for RL are often presented in terms of rewards, e.g. $r_t = -L_t$; however, losses are an equivalent representation that avoids complicating our equations with minus signs. Discounted future loss backpropagation through time\cite{werbos1990backpropagation} (BPTT) enables RNNs to be trained by gradient descent\cite{ruder2016overview}. However, losses for partial scan completions are not differentiable with respect to (w.r.t.) RNN actions, $(a_0,...,a_{T-1})$, controlling which path segments are sampled.


Many MDPs have losses that are not differentiable w.r.t. agent actions. Examples include agents directing their vision\cite{mnih2014recurrent, ba2014multiple}, managing resources\cite{alphastarblog}, and playing score-based computer games\cite{lillicrap2015continuous, heess2015memory}. Actors can be trained with non-differentiable losses by introducing a differentiable surrogate\cite{grabocka2019learning} or critic\cite{konda2000actor} to predict losses that can be backpropagated to actor parameters. Alternatively, non-differentiable losses can be backpropagated to agent parameters if actions are sampled from a differentiable probability distribution\cite{zhao2011analysis, mnih2014recurrent} as training losses given by products of losses and sampling probabilities are differentiable. There are also a variety of alternatives to gradient descent, such as simulated annealing\cite{rere2015simulated} and evolutionary algorithms\cite{young2015optimizing}, that do not require differentiable loss functions. Such alternatives can outperform gradient descent\cite{such2017deep}; however, they usually achieve similar or lower performance than gradient descent for deep ANN training.

\section{Training}

In this section, we outline our training environment, ANN architecture and learning policy. Our ANNs were developed in Python with TensorFlow\cite{abadi2016tensorflow}. Detailed architecture and learning policy is in supplementary information. In addition, source code and pretrained models are openly accessible from GitHub\cite{intelligent_scan_repo}, and training data is openly accessible\cite{ede2020warwick, warwickem!}.

\subsection{Environment}

To create partial scans from STEM images, an actor, $\mu$, infers action vectors, $\mu(h_t)$, based on a history, $h_t = (a_0, o_1^i, ..., a_{t-1}, o_t)$, of previous actions, $a$, and observations, $o$. To encourage exploration, $\mu(h_t)$ is rotated to $a_t$ by Ornstein-Uhlenbeck\cite{uhlenbeck1930theory} (OU) exploration noise\cite{plappert2017parameter}, $\epsilon_t$,
\begin{align}
a_t &= \begin{bmatrix}
\cos \epsilon_t & -\sin \epsilon_t \\
\sin \epsilon_t & \cos \epsilon_t 
\end{bmatrix} \mu(h_t) \\
\epsilon_t &= \theta (\epsilon_\text{avg}-\epsilon_{t-1}) + \sigma W
\end{align}
where we chose $\theta = 0.1$ to decay noise to $\epsilon_\text{avg}=0$, a scale factor, $\sigma=0.2$, to scale a standard normal variate, $W$, and start noise $\epsilon_0 = 0$. OU noise is linearly decayed to zero throughout training. Correlated OU exploration noise is recommended for continuous control tasks optimized by deep deterministic policy gradients\cite{lillicrap2015continuous} (DDPG) and recurrent deterministic policy gradients\cite{heess2015memory} (RDPG). Nevertheless, follow-up experiments with twin delayed deep deterministic policy gradients\cite{fujimoto2018addressing} (TD3) and distributed distributional deep deterministic policy gradients\cite{barth2018distributed} (D4PG) have found that uncorrelated Gaussian noise can produce similar results.

\begin{figure*}[tb!]
\centering
\includegraphics[width=0.8\textwidth]{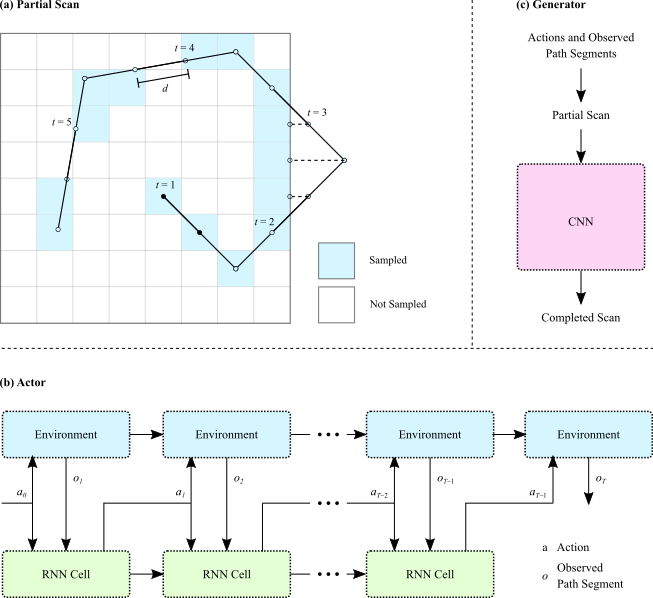}
\caption{ Simplified scan system. a) An example 8$\times$8 partial scan with $T=5$ straight path segments. Each segment in this example has 3 probing positions separated by $d=2^{1/2}$ px, and their starts are labelled by step numbers, $t$. Partial scans are selected from STEM images by sampling image pixels nearest probing positions, even if a nominal probing position is outside an imaging region. b) An actor RNN uses its previous state, action, and an observed path segment to choose the next action at each step. c) A partial scan constructed from actions and observed path segments is completed by a generator CNN. }
\label{fig:environment}
\end{figure*}

An action, $a_t$, is the direction to move to observe a path segment, $o_t$, from the position at the end of the previous path segment. Partial scans are constructed from complete histories of actions and observations, $h_T = (a_0, o_1, ..., a_{T-1}, o_T)$. A simplified partial scan is shown in figure~\ref{fig:environment}(a). In our experiments, partial scans, $s$, are constructed from $T=20$ straight path segments selected from 96$\times$96 STEM images. Each segment has 20 probing positions separated by $d=2^{1/2}$ px and positions can be outside an image. The pixels in the image nearest each probing position are sampled, so a separation of $d \ge 2^{1/2}$ simplied development by preventing successive probing positions in a segment from sampling the same pixel. A separation of $d < 2^{1/2}$ would allow a pixel to sampled more than once by moving diagonally, potentially incentivising orthogonal scan motion to sample more pixels.

Following our earlier work\cite{preprint+ede2019partial, preprint+ede2019deep, ede2020adaptive}, we select subsets of pixels from STEM images to create partial scans to train ANNs for compressed sensing. Selecting a subset of pixels is easier than preparing a large, carefully partitioned and representative dataset\cite{raschka2018model, roh2019survey} containing experimental partial scan and full image pairs, and selected pixels have realistic noise characteristics as they are from experimental images. However, selecting a subset of pixels does not account for probing location errors varying with scan shape\cite{sang2017dynamic}. We use a Warwick Electron Microscopy Dataset (WEMD) containing 19769 32-bit 96$\times$96 images cropped and downsampled from full images\cite{ede2020warwick, warwickem!}. Cropped images were blurred by a symmetric 5$\times$5 Gaussian kernel with a 2.5 px standard deviation to decrease any training loss variation due to varying noise characteristics. Finally, images, $I$, were linearly transformed to normalized images, $I_N$, with minimum and maximum values of $-1$ and $1$. To test performance, the 19769 images were split, without shuffling, into a training set containing 15815 images and a test set containing 3954 images.

\subsection{Architecture}

For training, our adaptive scan system consists of an actor, $\mu$, target actor, $\mu'$, critic, $Q$, target critic, $Q'$, and generator, $G$. Simplified actor and generator architecture is shown in figure~\ref{fig:environment}(b)-(c). To minimize latency, our actors and critics are computationally inexpensive deep LSTMs\cite{zaremba2014recurrent} with a depth of 2 and 256 hidden units. Our generator is a convolutional neural network\cite{mccann2017convolutional, krizhevsky2012imagenet} (CNN). A recurrent actor selects actions, $a_t$ and observes path segments, $o_t$, that are added to an experience replay\cite{zhang2017deeper}, $R$, containing $10^5$ complete histories of actions and observations. Partial scans, $s$, are constructed from histories sampled from the replay to train a generator to complete partial scans, $I_G^i = G(s^i)$. The actor and generator cooperate to minimize generator losses, $L_G$, and are the only networks needed for inference.

Generator losses are not differentiable w.r.t. actor actions used to construct partial scans i.e. $\partial L_G/\partial a_t = 0$. Following RDPG\cite{heess2015memory}, we therefore introduce recurrent critics to predict losses from actor actions and observations that can be backpropagated to actors for training by BPTT. Actor and critic RNNs have the same architecture, except actors have two outputs to parameterize actions whereas critics have one output to predict losses. Target networks\cite{lillicrap2015continuous, mnih2015human} use exponential moving averages of live actor and critic network parameters and are introduced to stabilize learning. For training by RDPG, live and target ANNs separately replay experiences. However, we propagate live RNN states to target RNNs at each step as a precaution against any cumulative divergence of target network behaviour from live network behaviour across multiple steps.

\begin{algorithm}
\caption{Cooperative recurrent deterministic policy gradients (CRDPG).}
\begin{algorithmic}
\STATE Initialize actor, $\mu$, critic, $Q$, and generator, $G$, networks with parameters $\omega$, $\theta$ and $\phi$, respectively.
\STATE Initialize target networks, $\mu'$ and $Q'$, with parameters $\omega' \leftarrow \omega$, $\theta' \leftarrow \theta$, respectively.
\STATE Initialize replay buffer, $R$.
\STATE Initialize average generator loss, $L_\text{avg}$.
\FOR{iteration $m = 1, M$}
  \STATE Initialize empty history, $h_0$.
  \FOR{step $t=1, T$}
    \STATE Make observation, $o_t$.
    \STATE $h_t \leftarrow h_{t-1}, a_{t-1}, o_t$ (append action and corresponding observation to history).
    \STATE Select action, $a_t$, by computing $\mu(h_t)$ and applying exploration noise, $\epsilon_t$.
  \ENDFOR
  \STATE Store the sequence $(a_0, o_1, ..., a_{T-1}, o_T)$ in $R$.
  \STATE Sample a minibatch of $N$ histories, $h_T^i = (a_0^i, o_1^i, ..., a_{T-1}^i, o_T^i)$, from $R$.
  \STATE Construct partial scans, $s^i$, from $h_T^i$.
  \STATE Use generator to complete partial scans, $I_G^i = G(s^i)$.
  \STATE Compute step losses, $(L_1^i, ..., L_T^i)$, from generator losses, $L_{G}^i$, and over edge losses, $E_t^i$,
  \begin{equation}
      L_t^i = E_t^i + \delta_{tT} \frac{\text{clip}(L_{G}^i)}{L_\text{avg}} \,,
  \end{equation}
  where the Kronecker delta, $\delta_{tT}$, is 1 if $t = T$ and 0 otherwise, and $\text{clip}(L_{G}^i)$ is the smaller of $L_{G}^i$ and three standard deviations above its running mean.
  \STATE Compute target values, $(y_1^i, ..., y_T^i)$, with target networks,
  \begin{equation}
      y_t^i = L_t^i + \gamma Q'(H_Q^i, a_t^i, o_{t+1}^i, \mu'(H_\mu^i, a_t^i, o_{t+1}^i)) \,,
  \end{equation}
  where $H_Q^i$ and $H_\mu^i$ are hidden states of live networks after computing $Q(h_t^i, a_t^i)$ and $\mu(h_t^i)$, respectively.
  \STATE Compute critic update (using BPTT),
  \begin{equation}
      \Delta \omega = \frac{1}{NT} \sum\limits_i^N \sum\limits_t^T 
      (y_t^i - Q(h_t^i, a_t^i)) \frac{\partial Q(h_t^i, a_t^i)}{\partial \omega} \,.
  \end{equation}
  \STATE Compute actor update (using BPTT),
  \begin{equation}
      \Delta \theta = \frac{1}{NT} \sum\limits_i^N \sum\limits_t^T
      \frac{\partial Q(h_t^i, a_t^i)}{\partial \mu(h_t^i)}
            \frac{\partial \mu(h_t^i)}{\partial \theta} \,.
  \end{equation}
  \STATE Compute generator update,
  \begin{equation}
      \Delta \phi = \frac{1}{N} \sum\limits_i^N \frac{\partial L_{G}^i}{\partial \phi} \,.
  \end{equation}
  \STATE Update the actor, critic, and generator by gradient descent.
  \STATE Update the target networks and average generator loss,
  \begin{align}
      \omega ' &\leftarrow \beta_{\omega} \omega ' + (1-\beta_{\omega}) \omega \,, \\
      \theta ' &\leftarrow \beta_{\theta} \theta ' + (1-\beta_{\theta}) \theta \,, \\
      L_\text{avg} &\leftarrow \beta_{L} L_\text{avg} + \frac{1-\beta_{L}}{N} \sum\limits_i^N (L_G^i) \,.
  \end{align}
\ENDFOR
\end{algorithmic}
\label{algorithm}
\end{algorithm}
 
\subsection{Learning Policy}

To train actors to cooperate with a generator to complete partial scans, we developed cooperative recurrent deterministic policy gradients (CRDPG, algorithm~\ref{algorithm}). This is an extension of RDPG to an actor that cooperates with another ANN to minimize its loss. We train our networks by ADAM\cite{kingma2014adam} optimized gradient descent for $M=10^6$ iterations with a batch size, $N=32$. We use constant learning rates $\eta_\mu = 0.0005$ and $\eta_Q = 0.0010$ for the actor and critic, respectively. For the generator, we use an initial learning rate $\eta_G = 0.0030$ with an exponential decay factor of $0.75^{5m/M}$ at iteration $m$. The exponential decay envelope is multiplied by a sawtooth cyclic learning rate\cite{smith2017cyclical} with a period of $2M/9$ that oscillates between 0.2 and 1.0. Training takes two days with an Intel i7-6700 CPU and an Nvidia GTX 1080 Ti GPU.

We augment training data by a factor of eight by applying a random combination of flips and 90$\degree$ rotations, mapping $s \to s'$ and $I_N \to I_N'$, similar to our earlier work\cite{ede2019improving, ede2020adaptive, preprint+ede2019partial, preprint+ede2019deep}. Our generator is trained to minimize mean squared errors (MSEs), 
\begin{equation}
    L_G = \text{MSE}(G(s'), I_N) \,,
\end{equation}
between scan completions, $G(s')$, and normalized target images, $I_N$. Generator losses decrease during training as the generator learns, and may vary due to loss spikes\cite{ede2020adaptive}, learning rate oscillations\cite{smith2017cyclical} or other training phenomena. Normalizing losses can improve RL\cite{van2016learning}, so we divide generator losses used for critic training by their running mean,
\begin{equation}
    L_\text{avg} \leftarrow \beta_{L} L_\text{avg} + \frac{1-\beta_{L}}{N} \sum\limits_i^N L_G \,,
\end{equation}
where we chose $\beta_{L} = 0.997$ and $L_\text{avg}$ is updated at each training iteration.

Heuristically, an optimal policy does not go over image edges as there is no information there in our training environment. To accelerate convergence, we therefore added a small loss penalty, $E_t=0.1$, at step $t$ if an action results in a probing position being over an image edge. The total loss at each step is
\begin{equation}\label{eqn:normalization}
    L_t = E_t + \delta_{tT} \frac{\text{clip}(L_G)}{L_\text{avg}} \,,
\end{equation}
where $\text{clip}(L_G)$ clips losses used for RL to three standard deviations above their running mean. This adaptive loss clipping is inspired by adaptive learning rate clipping\cite{ede2020adaptive} (ALRC) and reduces learning destabilization by high loss spikes. However, we expect that clipping normalized losses to a fixed threshold\cite{mnih2015human} would achieve similar results. The Kronecker delta, $\delta_{tT}$, in equation~\ref{eqn:normalization} is 1 if $t=T$ and 0 otherwise, so it only adds the generator loss at the final step, $T$. 

To estimate discounted future losses, $Q_t^\text{rl}$, for RL, we use a target actor and critic,
\begin{equation}\label{eqn:rl_loss}
    Q_t^\text{rl} =  L_t + \gamma Q'(H_Q, a_t, o_{t+1}, \mu'(H_\mu^i, a_t, o_{t+1})) \,,
\end{equation}
where we chose $\gamma=0.97$, and $H_Q^i$ and $H_\mu^i$ are hidden states of live networks after computing $Q(h_t^i, a_t^i)$ and $\mu(h_t^i)$, respectively. Target networks stabilize learning and decrease policy oscillations\cite{czarnecki2019distilling, lipton2016combating, wagner2011reinterpretation}. The critic is trained to minimize mean squared differences, $L_Q$, between predicted and target losses, and the actor is trained to minimize losses, $L_\mu$, predicted by the critic,
\begin{align}
    L_Q &= \frac{1}{2T} \sum\limits_{t=1}^T (y_t - Q(h_t, a_t))^2 \,, \\
    L_\mu &= \frac{1}{T} \sum\limits_{t=1}^T Q(h_t, a_t) \,.
\end{align} 
Our target actor and critic have trainable parameters $\omega'$ and $\theta'$, respectively, that track live parameters, $\omega$ and $\theta$, by soft updates\cite{lillicrap2015continuous},
\begin{align}
      \omega_m ' &= \beta_{\omega} \omega_{m-1} ' + (1-\beta_{\omega}) \omega_m \,,\\
      \theta_m ' &= \beta_{\theta} \theta_{m-1} ' + (1-\beta_{\theta}) \theta_m \,, 
\end{align}
where we chose $\beta_{\omega} = \beta_{\theta} = 0.9997$. We also investigated hard updates\cite{mnih2015human}, where target networks are periodically copied from live networks; however, we found that soft updates result in faster convergence and more stable training.

\begin{figure*}[tbh!]
\centering
\includegraphics[width=\textwidth]{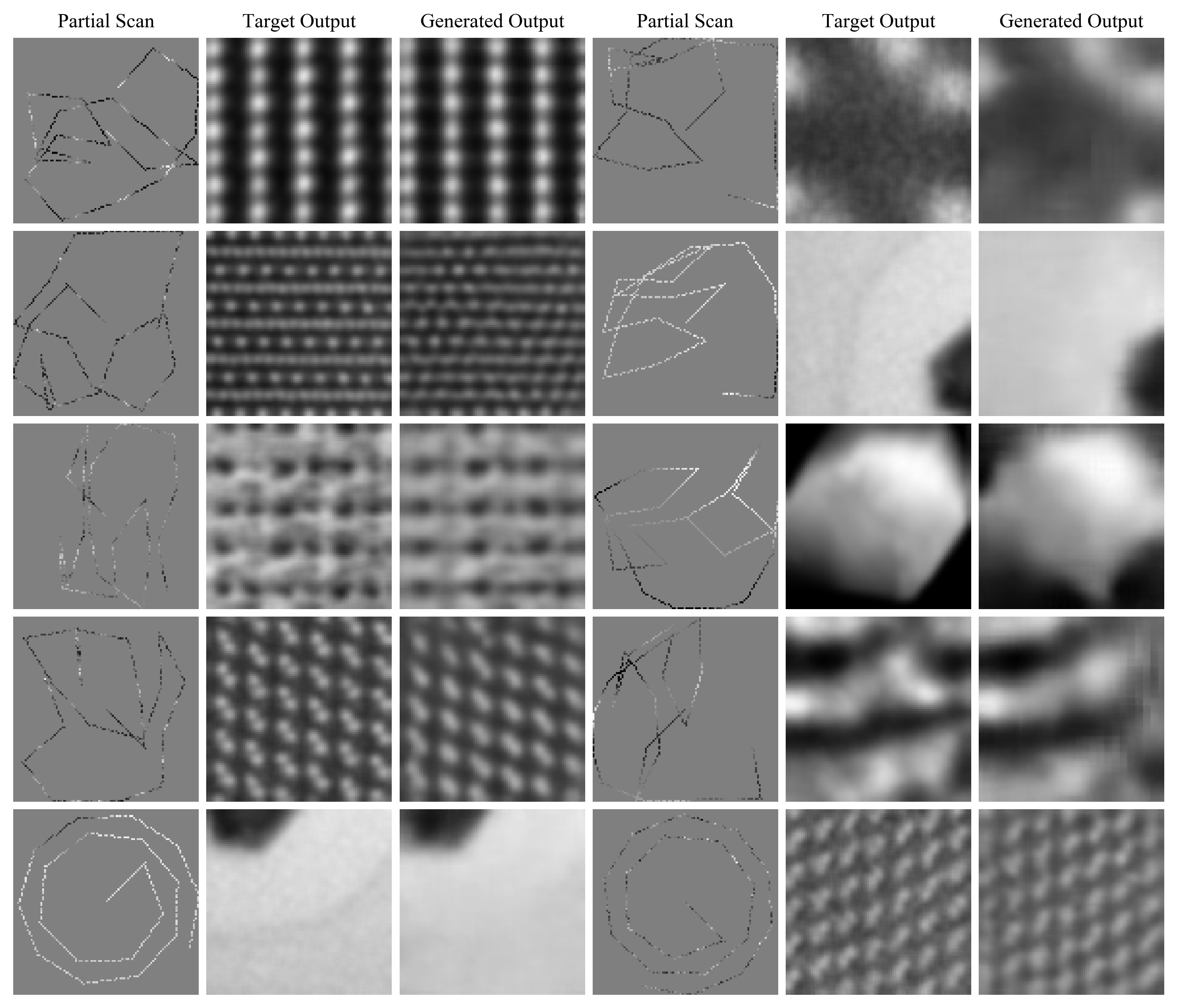}
\caption{ Examples of test set 1/23.04 px coverage partial scans, target outputs and generated partial scan completions for 96$\times$96 crops from STEM images. The top four rows show adaptive scans, and the bottom row shows spiral scans. Input partial scans are noisy, whereas target outputs are blurred. }
\label{fig:examples}
\end{figure*}

\begin{figure*}[tbh!]
\centering
\includegraphics[width=0.98\textwidth]{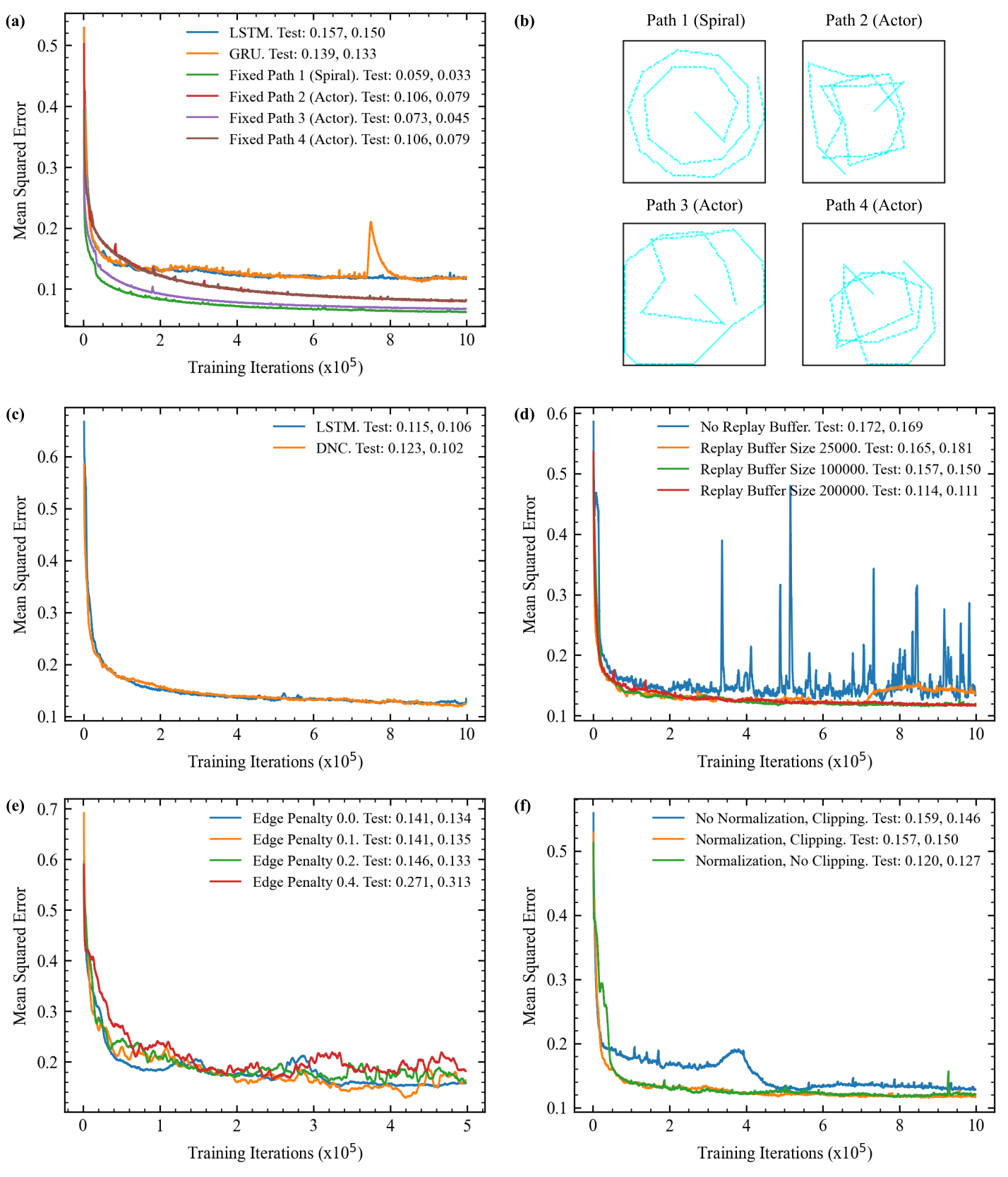}
\caption{ Learning curves for a)-b) adaptive scan paths chosen by an LSTM or GRU, and fixed spiral and other fixed paths, c) adaptive paths chosen by an LSTM or DNC, d) a range of replay buffer sizes, e) a range of penalties for trying to sample at probing positions over image edges, and f) with and without normalizing or clipping generator losses used for critic training. All learning curves are 2500 iteration boxcar averaged and results in different plots are not directly comparable due to varying experiment settings. Means and standard deviations of test set errors, \enquote{Test: Mean, Std Dev}, are at the ends of labels in graph legends. }
\label{fig:graph1}
\end{figure*}

\section{Experiments}

In this section, we present examples of adaptive partial scans and select learning curves for architecture and learning policy experiments. Examples of 1/23.04 px coverage partial scans, target outputs and generator completions are shown in figure~\ref{fig:examples} for 96$\times$96 crops from test set STEM images. They show both adaptive and spiral scans after flips and rotations to augment data for the generator. The first actions select a path segment from the middle of image in the direction of a corner. Actors then use the first and following observations to inform where to sample the remaining $T-1=19$ path segments. Actors adapt scan paths to specimens. For example, if an image contains regular atoms, an actor might cover a large area to see if there is a region where that changes. Alternatively, if an image contains a uniform region, actors, may explore near image edges and far away from the uniform region to find region boundaries.

The main limitation of our experiments is that generators trained to complete a variety of partial scan paths generated by an actor achieves lower performance than a generate trained to complete partial scans with a fixed path. For example, figure~\ref{fig:graph1}(a) shows that generators trained to cooperate with LSTM or GRU actors are outperformed by generators trained with fixed spiral or other scan paths shown in figure~\ref{fig:graph1}(b). Spiral paths outperform fixed scan paths; however, we emphasize that paths generated by actors are designed for individual training data, rather than all training data. Freezing actor training to prevent changes in actor policy does not result in clear improvements in generator performance. Consequently, we think that improvements to generator architecture or learning policy should be a starting point for further investigation. To find the best practical actor policy, we think that a generator trained for a variety of scan paths should achieve comparable performance to generators trained for single scan paths. 

We investigated a variety of popular RNN architectures to minimize inference time. Learning curves in figure~\ref{fig:graph1}(a) show that performance is similar for LSTMs and GRUs. GRUs require less computation. However, LSTM and GRU inference time is comparable and GRU training seems to be more prone to loss spikes, so LSTMs may be preferable. We also created a DNC by augmenting a deep LSTM with dynamic external memory. However, figure~\ref{fig:graph1}(c) shows that LSTM and DNC performance is similar, and inference time and computational requirements are much higher for our DNC. We tried to reduce computation and accelerate convergence by applying projection layers to LSTM hidden states\cite{jia2017long}. However, we found that performance decreased with decreasing projection layer size. 

Experienced replay buffers for RL often have heuristic sizes, such as $10^6$ examples. However, RL can be sensitive to replay buffer size\cite{zhang2017deeper}. Indeed, learning curves in figure~\ref{fig:graph1}(d) show that increasing buffer size improves learning stability and decreases test set errors. Increasing buffer size usually improves learning stability and decreases forgetting by exposing actors and critics to a higher variety of past policies. However, we expect that convergence would be slowed if the buffer became too large as increasing buffer size increases expected time before experiences with new policies are replayed. We also found that increasing buffer sized decreased the size of small loss oscillations\cite{czarnecki2019distilling, lipton2016combating, wagner2011reinterpretation}, which have a period near 2000 iterations. However, the size of loss oscillations does not appear to affect performance.

We found that initial convergence is usually delayed if a large portion of initial actions go outside the imaging region. This would often delay convergence by about $10^4$ iterations before OU noise led to the discovery of better exploration strategies away from image edges. Although $10^4$ iterations is only 1\% of our $10^6$ iteration learning policy, it often impaired development by delaying debugging or evaluation of changes to architecture and learning policy. Augmenting RL losses with subgoal-based heuristic rewards can accelerate convergence by making problems more tractable\cite{ng1999policy}. Thus, we added loss penalties if actors tried to go over image edges, which accelerated initial convergence. Learning curves in figure~\ref{fig:graph1}(e) show that over edge penalties at each step smaller than $E_t = 0.2$ have a similar effect on performance. Further, performance is lower for higher over edge penalties, $E_t \ge 0.2$. We also found that training is more stable if over edge penalties are added at individual steps, rather than propagated to past steps as part of a discounted future loss. 

Our actor, critic and generator are trained together. It follows that generator losses, which our critic learns to predict, decrease throughout training as generator performance improves. However, normalizing loss sizes usually improves RL\cite{van2016learning}, so we divide by their running means in equation~\ref{eqn:normalization}. Learning curves in figure~\ref{fig:graph1}(f) show that loss normalization improves learning stability and decreases final errors. Clipping training losses can improve RL\cite{mnih2015human}, so we clipped generator losses used for critic training to 3 standard deviations above their running means. We found that clipping increases test set errors, possibly because most training errors are in a similar regime. Thus, we expect that clipping may be more helpful for training with sparser scans as higher uncertainty may increase likelihood of unusually high generator losses.

\section{Discussion}

The main limitation of our adaptive scan system is that generator errors are much higher when a generator is trained for a variety of scan paths than when it is trained for a single scan path. However, we expect that generator performance for a variety of scans could be improved to match performance for single scans by developing a larger neural network with a better learning policy. To train actors to cooperate with generators, we developed CRDPG. This is an extension of RDPG\cite{heess2015memory}, and RDPG is based on DDPG\cite{lillicrap2015continuous}. Alternatives to DDPG, such as TD3\cite{fujimoto2018addressing} and D4PG\cite{barth2018distributed}, arguably achieve higher performance, so we expect that they could form the basis of a future training algorithm. Further, we expect that architecture and learning policy could be improved by AdaNet\cite{weill2019adanet}, Ludwig\cite{molino2019ludwig}, or other automatic machine learning\cite{he2019automl, malekhosseini2019modeling, jaafra2019reinforcement, elsken2018neural, waring2020automated} (AutoML) algorithms as AutoML can often match or surpass the performance of human developers\cite{hanussek2020can, zoph2018learning}. Finally, test set losses for a variety of scans appear to be decreasing at the end of training, so we expect that performance could be improved by increasing training iterations.

After generator performance is improved, we expect the main limitation of our adaptive scan system to be distortions caused by probing position errors. Errors usually depend on scan path shape\cite{sang2017dynamic} and accumulate for each path segment. Non-linear scan distortions can be corrected by comparing pairs of orthogonal raster scans\cite{ophus2016correcting, ning2018scanning}, and we expect this method can be extended to partial scans. However, orthogonal scanning would complicate measurement by limiting scan paths to two half scans to avoid doubling electron dose on beam-sensitive materials. Instead, we propose that a cyclic generator\cite{zhu2017unpaired} could be trained to correct scan distortions and provide a detailed method as supplementary information\cite{ede2020adaptive_scans_supplementary}. Another limitation is that our generators do not learn to correct STEM noise\cite{seki2018theoretical}. However, we expect that generators can learn to remove noise, for example, from single noisy examples\cite{laine2019high} or by supervised learning\cite{ede2019improving}.

To simplify our preliminary investigation, our scan system samples straight path segments and cannot go outside a specified imaging region. However, actors could learn to output actions with additional degrees of freedom to describe curves, multiple successive path segments, or sequences of non-contiguous probing positions. Similarly, additional restrictions could be applied to actions. For example, actions could be restricted to avoid actions that cause high probing position errors. Training environments could also be modified to allow actors to sample pixels over image edges by loading images larger than partial scan regions. In practice, actors can sample outside a scan region and being able to access extra information outside an imaging region could improve performance. However, using larger images may slow development by increasing data loading and processing times.

Not all scan systems support non-raster scan paths. However, many scan controllers can be augmented with an FPGA to enable custom scan paths\cite{sang2017dynamic, Sang2017a}. Recent versions of Gatan DigitalMicrograph support Python\cite{miller2019real}, so our ANNs can be readily integrated into existing scan systems. Alternatively, an actor could be synthesized on a scan-controlling FPGA\cite{noronha2018leflow, ruan2019reinforcement} to minimize inference time. There could be hundreds of path segments in a partial scan, so computationally lightweight and parallelizable actors are essential to minimize scan time. We have therefore developed actors based computationally inexpensive RNNs, which can remember state information to inform future decisions. Another approach is to update a partial scan at each step to be input to feedforward neural network (FNN), such as a CNN, to decide actions. However, we expect that FNNs are less practical than RNNs as FNNs may require additional computation to reprocess all past states at each step.

\section{Conclusions}

Our initial investigation demonstrates that actor RNNs can be trained by RL to direct piecewise adaption of contiguous scans to specimens for compressed sensing. We introduce CRDPG to train an RNN to cooperate with a CNN to complete STEM images from partial scans and present our learning policy, experiments, and example applications. After further development, we expect that adaptive scans will become the most effective approach to decrease electron beam damage and scan time with minimal information loss. Static sampling strategies are a subset of possible dynamic sampling strategies, so the performance of static sampling can always be matched by or outperformed by dynamic sampling. Further, we expect that adaptive scan systems can be developed for most areas of science and technology, including for the reduction of medical radiation. To encourage further investigation, our source code, pretrained models, and training data is openly accessible.

\section{Supplementary Information}

Supplementary information is openly accessible at \url{https://doi.org/10.5281/zenodo.4384708}. It presents detailed ANN architecture, additional experiments and example scans, and a new method to correct partial scan distortions.

\section*{Data Availability}

The data that support the findings of this study are openly available. 

\section*{Acknowledgements}

Thanks go to Jasmine Clayton, Abdul Mohammed, and Jeremy Sloan for internal review. The author acknowledges funding from EPSRC grant EP/N035437/1 and EPSRC Studentship 1917382.

\section*{Competing Interests}

The author declares no competing interests.

\bibliography{bibliography}

\end{document}


\flushbottom
\maketitle

\begin{figure*}[tbh!]
\centering
\includegraphics[width=\textwidth]{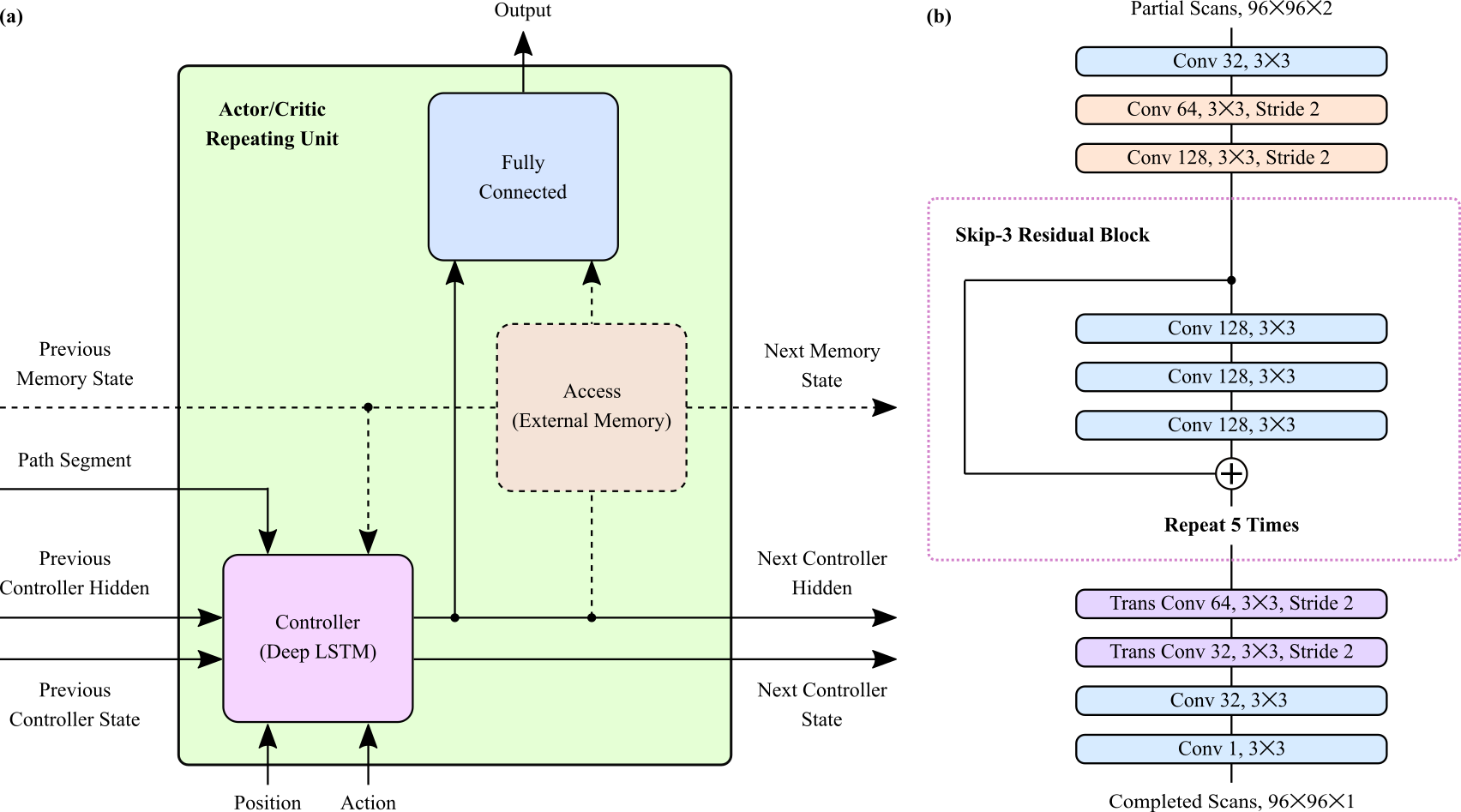}
\caption{ Actor, critic and generator architecture. a) An actor outputs action vectors whereas a critic predicts losses. Dashed lines are for extra components in a DNC. b) A convolutional generator completes partial scans. }
\label{fig:components}
\end{figure*}

\section{Detailed Architecture}

Detailed actor, critic and generator architecture is shown in figure~\ref{fig:components}. Actors and critics have almost identical architecture, except actor fully connected layers output action vectors whereas critic fully connected layers output predicted losses. In most of our experiments, actors and critics are deep LSTMs\cite{zaremba2014recurrent}. However, we also augment deep LSTMs with dynamic external memory to create DNCs\cite{graves2016hybrid} in some of our experiments. Configuration details of actor and critic components shown in figure~\ref{fig:components}(a) follow.

\vspace{\extraspace}
\noindent \textbf{Controller (Deep LSTM):} A two-layer deep LSTM with 256 hidden units in each layer. To reduce signal attenuation, we add skip connections from inputs to the second LSTM layer and from the first LSTM layer to outputs. Weights are initialized from truncated normal distributions and biases are zero initialized. In addition, we add a bias of 1 to the forget gate to reduce forgetting at the start of training\cite{jozefowicz2015empirical}. Initial LSTM cell and hidden states are initialized with trainable variables\cite{pitis2016nonzero}.

\vspace{\extraspace}
\noindent \textbf{Access (External Memory):} Our DNC implementation is adapted from Google Deepmind's\cite{graves2016hybrid, dnc_repo}. We use 4 read heads and 1 write head to control access to dynamic external memory, which has 16 slots with a word size of 64.

\vspace{\extraspace}
\noindent \textbf{Fully Connected:} A dense layer linearly connects inputs to outputs. Weights are initialized from a truncated normal distribution and there are no biases.

\vspace{\extraspace}
\noindent \textbf{Conv \textit{d}, \textit{w}x\textit{w}, Stride, \textit{x}:} Convolutional layer with a square kernel of width, $w$, that outputs $d$ feature channels. If the stride is specified, convolutions are only applied to every $x$th spatial element of their input, rather than to every element. Striding is not applied depthwise.

\vspace{\extraspace}
\noindent \textbf{Trans Conv \textit{d}, \textit{w}x\textit{w}, Stride, \textit{x}:} Transpositional convolutional layer with a square kernel of width, $w$, that outputs $d$ feature channels. If the stride is specified, convolutions are only applied to every $x$th spatial element of their input, rather than to every element. Striding is not applied depthwise.

\vspace{\extraspace}
\noindent \textbf{\circled{+}:} Circled plus signs indicate residual connections where incoming tensors are added together. Residuals help reduce signal attenuation and allow a network to learn perturbative transformations more easily.

\vspace{\extraspace}
\noindent The actor and critic cooperate with a convolutional generator, shown in figure~\ref{fig:components}(b), to complete partial scans. Our generator is constructed from convolutional layers\cite{dumoulin2016guide} and skip-3 residual blocks\cite{he2015deep}. Each convolutional layer is followed by ReLU\cite{nair2010rectified} activation then batch normalization\cite{ioffe2015batch}, and residual connections are added between activation and batch normalization. The convolutional weights are Xavier\cite{glorot2010understanding} initialized and biases are zero initialized.

\begin{figure*}[tbh!]
\centering
\includegraphics[width=0.98\textwidth]{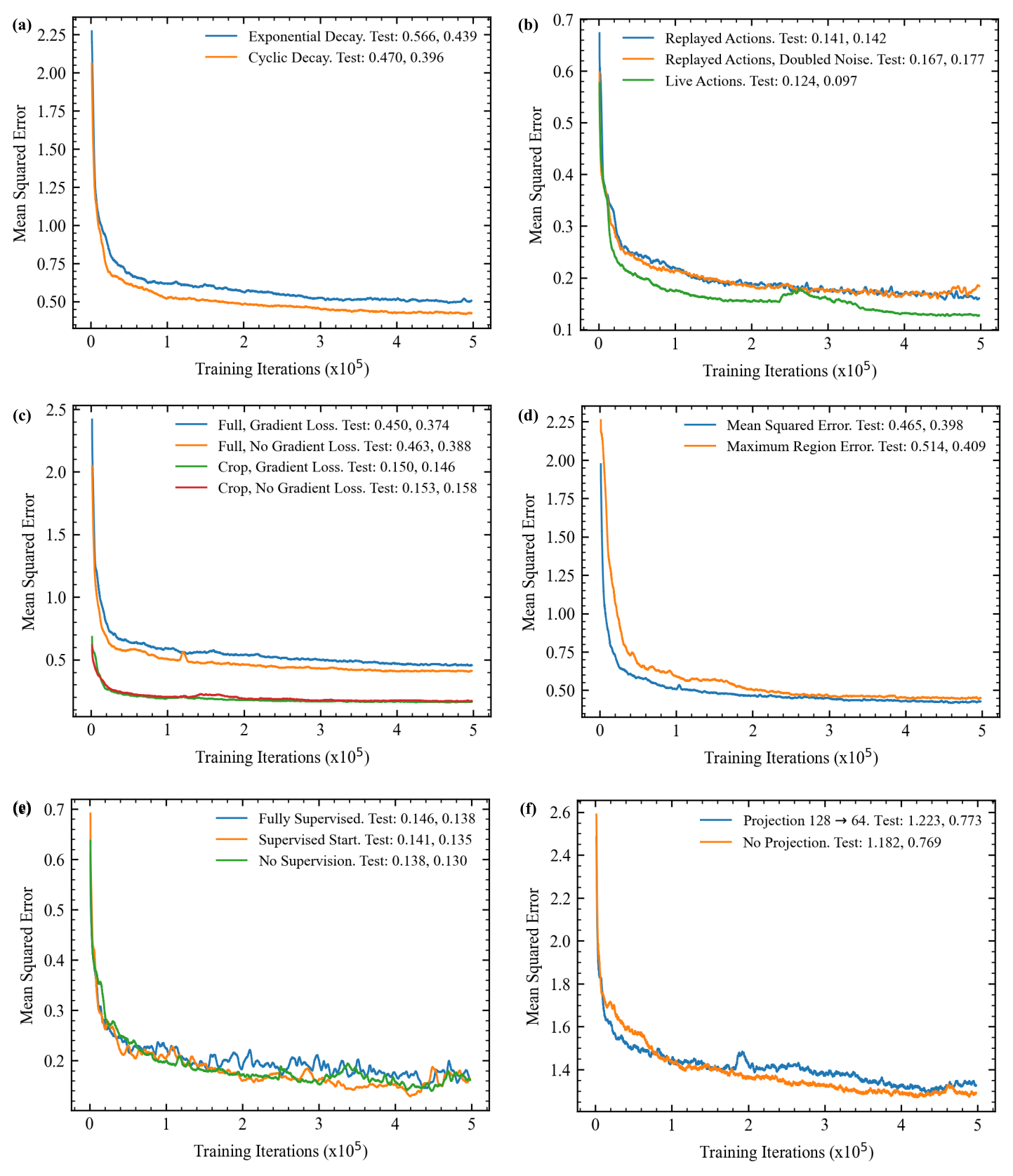}
\caption{ Learning curves for a) exponentially decayed and exponentially decayed cyclic learning rate schedules, b) actor training with differentiation w.r.t. live or replayed actions, c) images downsampled or cropped from full images to 96$\times$96 with and without additional Sobel losses, d) mean squared error and maximum regional mean squared error loss functions, e) supervision throughout training, supervision only at the start, and no supervision, and f) projection from 128 to 64 hidden units or no projection. All learning curves are 2500 iteration boxcar averaged, and results in different plots are not directly comparable due to varying experiment settings. Means and standard deviations of test set errors, \enquote{Test: Mean, Std Dev}, are at the ends of graph labels. }
\label{fig:graph2}
\end{figure*}

\begin{figure}
\centering 
\includegraphics[width=\columnwidth]{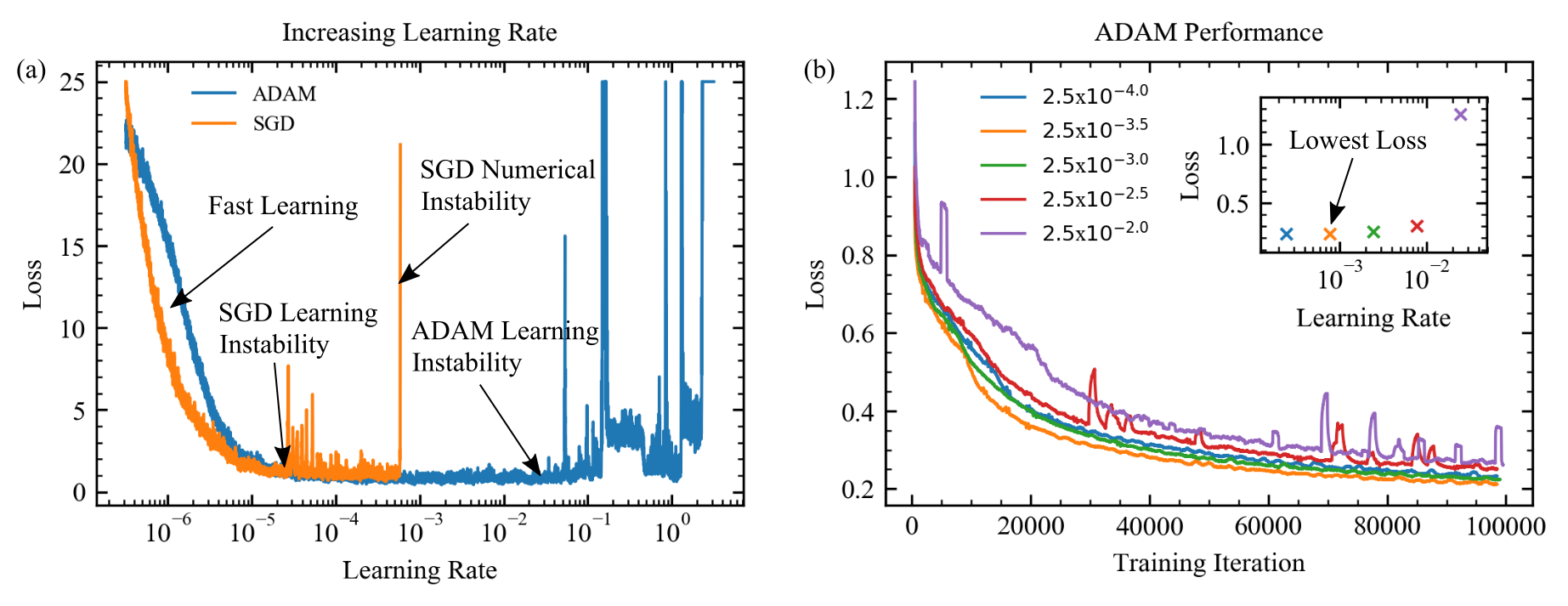}
\caption{ Learning rate optimization. a) Learning rates are increased from $10^{-6.5}$ to $10^{0.5}$ for ADAM and SGD optimization. At the start, convergence is fast for both optimizers. Learning with SGD becomes unstable at learning rates around 2.2$\times$10$^{-5}$, and numerically unstable near 5.8$\times$10$^{-4}$, whereas ADAM becomes unstable around 2.5$\times$10$^{-2}$. b) Training with ADAM optimization for learning rates listed in the legend. Learning is visibly unstable at learning rates of 2.5$\times$10$^{-2.5}$ and 2.5$\times$10$^{-2}$, and the lowest inset validation loss is for a learning rate of 2.5$\times$10$^{-3.5}$. Learning curves in (b) are 1000 iteration boxcar averaged. Means and standard deviations of test set errors, \enquote{Test: Mean, Std Dev}, are at the ends of graph labels. }
\label{fig:learning_rate_optimization}
\end{figure}

\section{Additional Regularization}

We apply L2 regularization\cite{kukavcka2017regularization} to decay generator parameters by a factor, $\beta = 0.99999$, at each training iteration. This decay rate is heuristic and the L2 regularization is primarily a precaution against overfitting. Further, adding L2 regularization did not have a noticeable effect on performance. We also investigated gradient clipping\cite{zhang2019gradient, gorbunov2020stochastic, chen2020understanding, menon2019can} to a range of static and dynamic thresholds for actor and critic training. However, we found that gradient clipping decreases convergence if clipping thresholds are too small and otherwise does not have a noticeable effect. 

\section{Additional Experiments}

This section present additional learning curves for architecture and learning policy experiments in figure~\ref{fig:graph2}. For example, learning curves in figure~\ref{fig:graph2}(a) show that generator training with an exponentially decayed cyclic learning rate\cite{smith2017cyclical} results in faster convergence and lower final errors than just using an exponentially decayed learning rate. We were concerned that a cyclic learning rate might cause generator loss oscillations if the learning rate oscillated too high. Indeed, our investigation of loss normalization was, in part, to prevent potential generator loss oscillations from destabilizing critic training. However, our learning policy results in generator losses that steadily decay throughout training.

To train actors by BPTT, we differentiate losses predicted by critics w.r.t. actor parameters by the chain rule,
\begin{equation}
  \Delta \theta = \frac{1}{NT} \sum\limits_i^N \sum\limits_t^T
  \frac{\partial Q(h_t^i, a_t^i)}{\partial \theta} = \frac{1}{NT} \sum\limits_i^N \sum\limits_t^T
  \frac{\partial Q(h_t^i, a_t^i)}{\partial \mu(h_t^i)} \frac{\partial \mu(h_t^i)}{\partial \theta} \,.
\end{equation}
An alternative approach is to replace $\partial Q(h_t^i, a_t^i)/\partial \mu(h_t^i)$ with a derivative w.r.t. replayed actions, $\partial Q(h_t^i, a_t^i)/\partial a_t^i$. This is equivalent to adding noise, $\text{stop\_gradient}(a_t^i-\mu(h_t^i))$, to an actor action, $\mu(h_t^i)$, where $\text{stop\_gradient}(x)$ is a function that stops gradient backpropagation to $x$. However, learning curves in figure~\ref{fig:graph2}(b) show that differentiation w.r.t. live actor actions results in faster convergence to lower losses. Results for $\partial Q(h_t^i, a_t^i)/\partial a_t^i$ are similar if OU exploration noise is doubled.

Most STEM signals are imaged at several times their Nyquist rates\cite{ede2020warwick}. To investigate adaptive STEM performance on signals imaged close to their Nyquist rates, we downsampled STEM images to 96$\times$96. Learning curves in figure~\ref{fig:graph2}(c) show that losses are lower for oversampled STEM crops. Following, we investigated if MSEs vary for training with different loss metrics by adding a Sobel loss, $\lambda_S L_S$, to generator losses. Our Sobel loss is
\begin{equation}
    L_S = \text{MSE}(S(G(s)), S(I_N)) \,,
\end{equation}
where $S(x)$ computes a channelwise concatenation of horizontal and vertical Sobel derivatives\cite{vairalkar2012edge} of x, and we chose $\lambda_S = 0.1$ to weight the contribution of $L_S$ to the total generator loss, $L_G + \lambda_S L_S$. Learning curves in figure~\ref{fig:graph2}(c) show that Sobel losses do not decrease training MSEs for STEM crops. However, Sobel losses decrease MSEs for downsampled STEM images. This motivates the exploration of alternative loss functions\cite{zhao2015loss} to further improve performance. For example, our earlier work shows that generator training as part of a generative adversarial network\cite{gui2020review, saxena2020generative, pan2019recent, wang2019generative} (GAN) can improve STEM image realism\cite{preprint+ede2019partial}. Similarly, we expect that generated image realism could be improved by training generators with perceptual losses\cite{grund2020improving}.

After we found that adding a Sobel loss can decrease MSEs, we also experimented with other loss functions, such as the maximum MSE of 5$\times5$ regions. Learning curves in figure~\ref{fig:graph2}(d) show that MSEs result in faster convergence than maximum region losses; however, both loss functions result in similar final MSEs. We expect that MSEs calculated with every output pixel result in faster convergence than maximum region errors as more pixels inform gradient calculations. In any case, we expect that a better approach to minimize maximum errors is to use a higher order loss function, such as mean quartic errors. If training with a higher-order loss function is unstable, it might be stabilized by adaptive learning rate clipping\cite{ede2020adaptive}.

Target losses can be directly computed with Bellman's equation, rather than with target networks. We refer to such directly computed target losses as \enquote{supervised} losses,
\begin{equation}\label{eqn:super_loss}
    Q_t^\text{super} =  \sum\limits_{t'=t}^T \gamma^{t'-t} L_{t'} \,,
\end{equation}
where where $\gamma \in [0,1)$ discounts future step losses, $L_t$. Learning curves for full supervision, supervision linearly decayed to zero in the first $10^5$ iterations, and no supervision are shown in figure~\ref{fig:graph2}(e). Overall, final errors are similar for training with and without supervision. However, we find that learning is usually more stable without supervised losses. As a result, we do not recommend using supervised losses.

To accelerate convergence and decrease computation, an LSTM with $n_h$ hidden units can be augmented by a linear projection layer with $n_p < 3n_h/4$ units\cite{jia2017long}. Learning curves in figure~\ref{fig:graph2}(f) are for $n_h = 128$ and compare training with a projection to $n_p = 64$ units and no projection. Adding a projecting layer increases the initial rate of convergence; however, it also increases final losses. Further, we found that training becomes increasingly prone to instability as $n_p$ is decreased. As a result, we do not use projection layers in our actor or critic networks.

Generator learning rate optimization is shown in figure~\ref{fig:learning_rate_optimization}. To find the best initial learning rate for ADAM optimization, we increased the learning rate until training became unstable, as shown in figure~\ref{fig:learning_rate_optimization}(a). We performed the learning rate sweep over $10^{4}$ iterations to avoid results being complicated by losses rapidly decreasing in the first couple of thousand. The best learning rate was then selected by training for $10^5$ iterations with learning rates within a factor of 10 from a learning rate 10$\times$ lower than where training became unstable, as shown in figure~\ref{fig:learning_rate_optimization}(b). We performed initial learning rate sweeps in figure~\ref{fig:learning_rate_optimization}(a) for both ADAM and stochastic gradient descent\cite{ruder2016overview} (SGD) optimization. We chose ADAM as it is less sensitive to hyperparameter choices than SGD and because ADAM is recommended in the RDPG paper\cite{heess2015memory}.

\section{Test Set Errors}

Test set errors are computed for 3954 test set images. Most test set errors are similar to or slightly higher than training set errors. However, training with fixed paths, which is shown in figure~{\maingraphnum}(a) of the main article, results in high divergence of test and training set errors. We attribute this divergence to the generator overfitting to complete large regions that are not covered by fixed scan paths. In comparison, our learning policy was optimized for training with a variety of adaptive scan paths where overfitting is minimal. After all $10^6$ training iterations, means and standard deviations (mean, std dev) of test set errors for fixed paths 2, 3 and 4 are (0.170, 0.182), (0.135, 0.133), and (0.171, 0.184). Instead, we report lower test set errors of (0.106, 0.090), (0.073, 0.045), and (0.106. 0.090), respectively, at $5 \times 10^5$ training iterations, which correspond to early stopping\cite{li2020gradient, flynn2020bounding}. All other test set errors were computed after final training iterations.

\section{Distortion Correction}

A limitation of partial STEM is that images are usually distorted by probing position errors, which vary with scan path shape\cite{sang2017dynamic}. Distortions in raster scans can be corrected by comparing series of images\cite{zhang2017joint, jin2015correction}. However, distortion correction of adaptive scans is complicated by more complicated scan path shapes and microscope-specific actor command execution characteristics. We expect that command execution characteristics are almost static. Thus, it follows that there is a bijective mapping between probing locations in distorted adaptive partial scans and raster scans. Subsequently, we propose that distortions could be corrected by a cyclic generative adversarial network\cite{zhu2017unpaired} (GAN). To be clear, this section outlines a possible starting point for future research that can be refined or improved upon. The method's main limitation is that the cyclic GAN would need to be trained or fine-tuned for individual scan systems.

Let $I_\text{partial}$ and $I_\text{raster}$ be unpaired partial scans and raster scans, respectively. A binary mask, $M$, can be constructed to be 1 at nominal probing positions in $I_\text{partial}$ and 0 elsewhere. We introduce generators $G_{p \rightarrow r}(I_\text{partial})$ and $G_{r \rightarrow p}(I_\text{raster}, M)$ to map from partial scans to raster scans and from raster scans to partial scans, respectively. A mask must be input to the partial scan generator for it to output a partial scan with a realistic distortion field as distortions depend on scan path shape\cite{sang2017dynamic}. Finally, we introduce discriminators $D_\text{partial}$ and $D_\text{raster}$ are trained to distinguish between real and generated partial scans and raster scans, respectively, and predict losses that can be used to train generators to create realistic images. In short, partial scans could be mapped to raster scans by minimizing 
\begin{align}
    L_{p \rightarrow r}^\text{GAN} &= D_\text{raster}(G_{p \rightarrow r}(I_\text{partial})) \,, \\
    L_{r \rightarrow p}^\text{GAN} &= D_\text{partial}(MG_{r \rightarrow p}(I_\text{raster}, M)) \,, \\
    L_{r \rightarrow p}^\text{cycle} &= \text{MSE}(MG_{r \rightarrow p}(G_{p \rightarrow r}(I_\text{partial}), M), I_\text{partial}) \,, \\
    L_{p \rightarrow r}^\text{cycle} &= \text{MSE}(G_{p \rightarrow r}(MG_{r \rightarrow p}(I_\text{raster}, M)), I_\text{raster}) \,, \\
    L_{p \rightarrow r} &= L_{p \rightarrow r}^\text{GAN} + b L_{r \rightarrow p}^\text{cycle} \,, \\
    L_{r \rightarrow p} &= L_{r \rightarrow p}^\text{GAN} + b L_{p \rightarrow r}^\text{cycle} \,,
\end{align}
where $L_{p \rightarrow r}$ and $L_{p \rightarrow r}$ are total losses to optimize $G_{p \rightarrow r}$ and $G_{p \rightarrow r}$, respectively. A scalar, $b$, balances adversarial and cycle-consistency losses.

\begin{figure*}[tbp!]
\centering
\includegraphics[width=0.95\textwidth]{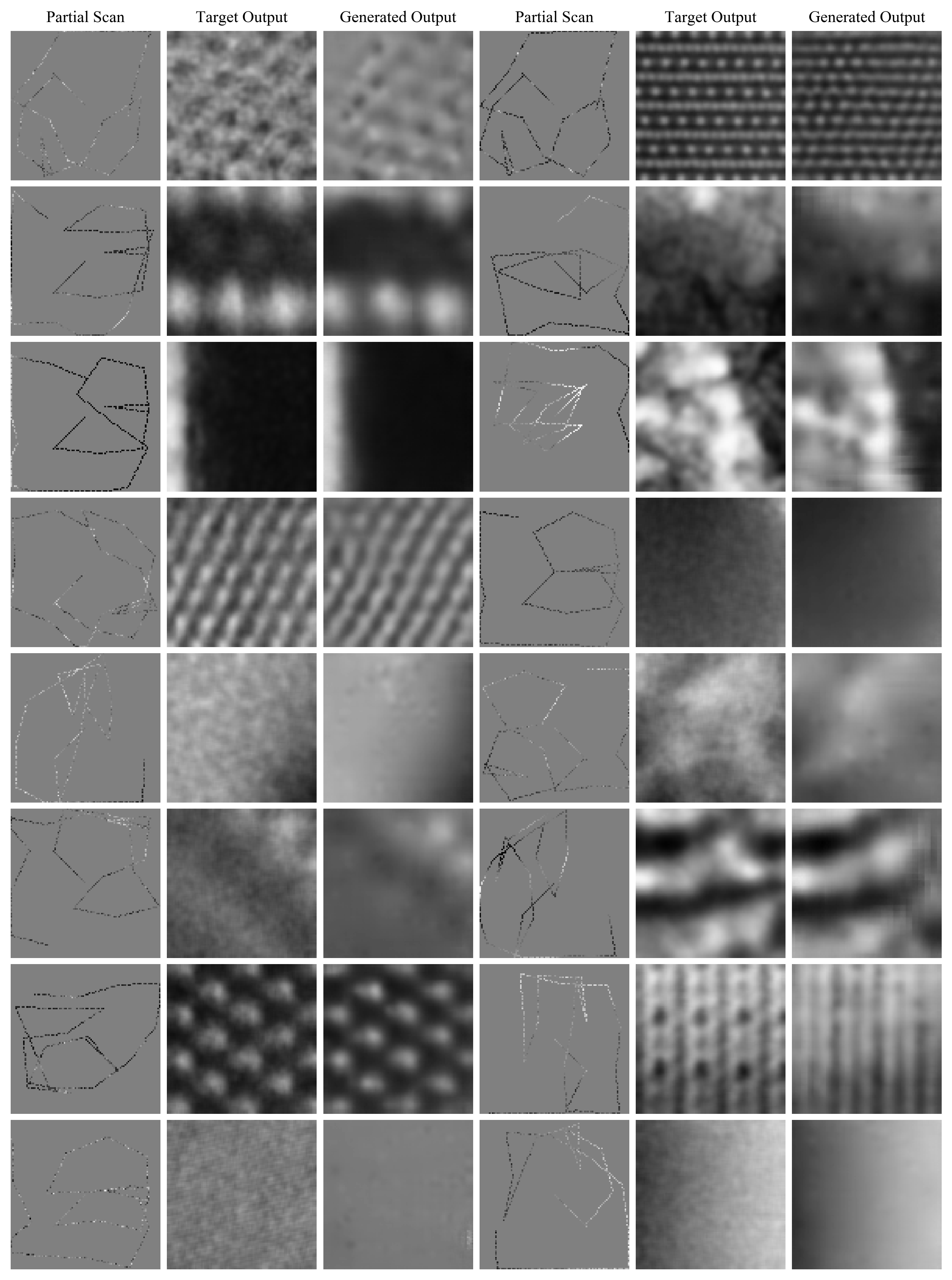}
\caption{ Test set 1/23.04 px coverage adaptive partial scans, target outputs, and generated partial scan completions for 96$\times$96 crops from STEM images. }
\label{fig:examples-2}
\end{figure*}

\begin{figure*}[tbp!]
\centering
\includegraphics[width=0.95\textwidth]{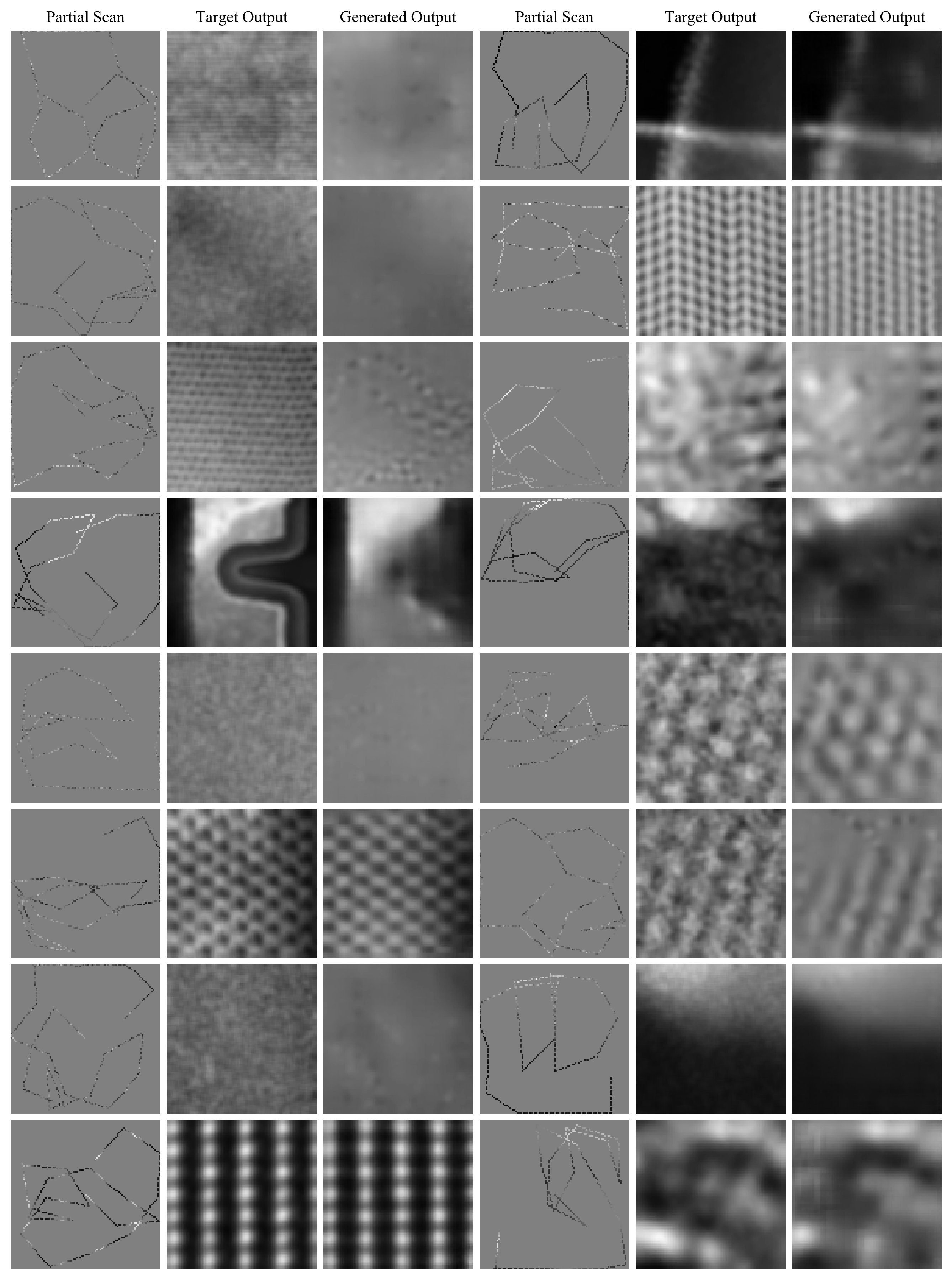}
\caption{ Test set 1/23.04 px coverage adaptive partial scans, target outputs, and generated partial scan completions for 96$\times$96 crops from STEM images. }
\label{fig:examples-1}
\end{figure*}

\begin{figure*}[tbp!]
\centering
\includegraphics[width=0.95\textwidth]{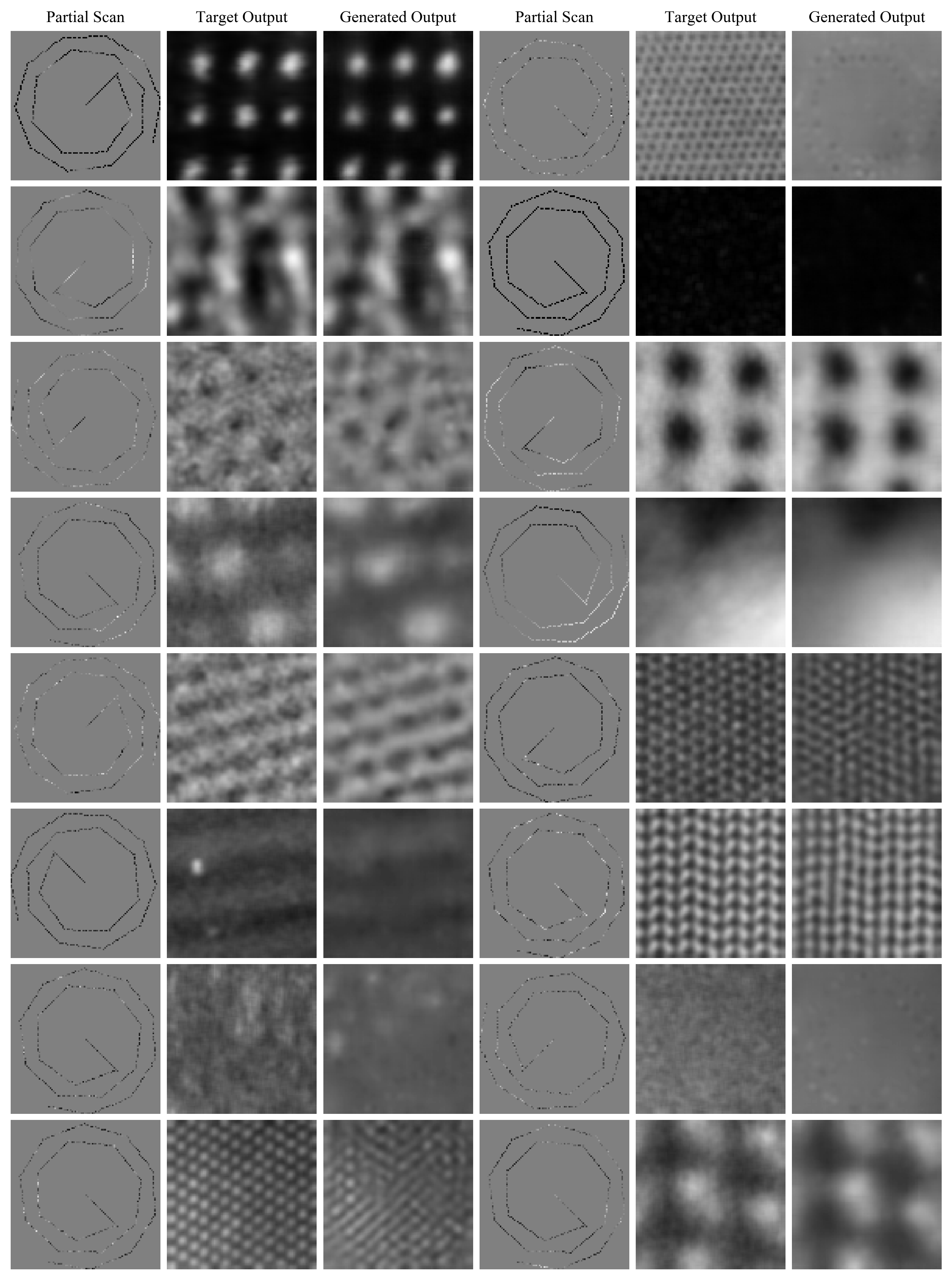}
\caption{ Test set 1/23.04 px coverage spiral partial scans, target outputs, and generated partial scan completions for 96$\times$96 crops from STEM images. }
\label{fig:examples-3}
\end{figure*}

\section{Additional Examples}

Additional sheets of test set adaptive scans are shown in figure~\ref{fig:examples-2} and figure~\ref{fig:examples-1}. In addition, a sheet of test set spiral scans is shown in figure~\ref{fig:examples-3}. Target outputs were low-pass filtered by a 5$\times$5 symmetric Gaussian kernel with a 2.5 px standard deviation to suppress high-frequency noise.

\clearpage
\bibliography{bibliography}